\documentclass{article}
\usepackage{times}
\usepackage{cite}
\usepackage{amsmath}
\usepackage[figurename=Figure]{caption}
\usepackage{graphicx}
\usepackage{latexsym}
\usepackage{subfig}
\usepackage{pstricks}
\usepackage{setspace}
\usepackage{tikz}
\usepackage{epstopdf}

\begin{document}
\title{A Survey of Brain Inspired Technologies for Engineering}
\author{
Jarryd Son \\
Electrical Engineering Department \\
University of Cape Town, South Africa\\
Email: jdsonza@gmail.com
\and
Amit Kumar Mishra \\
Electrical Engineering Department \\
University of Cape Town, South Africa\\
Email: akmishra@ieee.org
}

\maketitle
\begin{abstract}
Cognitive engineering is a multi-disciplinary field and hence it is difficult to find a review article consolidating the leading developments in the field. The incredible pace at which technology is advancing pushes the boundaries of what is achievable in cognitive engineering. There are also differing approaches to cognitive engineering brought about from the multi-disciplinary nature of the field and the vastness of possible applications. Thus research communities require more frequent reviews to keep up to date with the latest trends. In this paper we shall discuss some of the approaches to cognitive engineering holistically to clarify the reasoning behind the different approaches and to highlight their strengths and weaknesses. We shall then show how developments from seemingly disjointed views could be integrated to achieve the same goal of creating cognitive machines. By reviewing the major contributions in the different fields and showing the potential for a combined approach, this work intends to assist the research community in devising more unified methods and techniques for developing cognitive machines.

\emph{\textbf{artificial intelligence, cognitive architecture, bio-inspired}}
\end{abstract}
\maketitle
\section{Introduction}
The functioning of the brain has intrigued researchers since the beginning of scientific endeavours. The introduction of computers saw the advent of exciting developments which has culminated in the development of the new discipline of artificial intelligence (AI). Within the field of AI there has been a divided opinion on what the best approach is to create cognitive machines \cite{intelligence2003modern}. This division is based primarily on how information is claimed to be processed in the brain \cite{smolensky1987, HARNAD1990335}. On one side is the symbolic approach and the other is the sub-symbolic approach. 

Symbolic approaches such as cognitive architectures have a long history in AI and their developments have been devoted towards creating computational models that formalize the structure of the human brain \cite{langley2009cognitive}. Cognitive architectures such as Soar \cite{laird1987soar} and ACT-R \cite{anderson2013architecture} have been under development for many decades and have been successfully applied in various studies.

There have been arguments against the use of symbol systems because they oversimplify the underlying mechanisms required for cognition \cite{smolensky1987}. The alternative is to mimic the biology of brains, which gives rise to the sub-symbolic, connectionist approach. Connectionist approaches such as artificial neural networks (ANN's) have been through several generations of major developments and have become the leading technology in AI in recent years \cite{lowry2016visual}.   

The work discussed in this paper is focussed on how the mammalian brain can provide insights into creating appropriate models for designing cognitive machines. Furthermore, this work suggests unifying the differing approaches to create a holistic model rather than narrowing in on specific features. This work contributes to the engineering community by exploring useful technologies that could assist in creating more intelligent machines. Machine learning and AI has had a strong focus in tasks such as image recognition, language translation and financial analytics, however applications of such technology for machines that interact with the physical world has been less prominent. 

The paper is organized as follows: Section II looks at symbolic AI approaches, Section III looks at the increasingly popular connectionist approaches, Section IV identifies the advantageous qualities of a hybrid approach, Section V reviews advances in creating specialised hardware that mimics biology of the brain, and lastly the paper concludes in Section VI.  

\section{Symbolic Approaches}
Robotics is often cited as a field where AI and machine learning technology can be used, however many of the attempts focus on perceptual systems and ignore high-level cognitive capabilities \cite{hanford2011cognitive}. Much of the early success in achieving such capabilities was through the use of symbolic AI systems. AI pioneers, Alan Newell and Herbert A. Simon formulated the physical symbol system hypothesis that claims that, ``a physical symbol system has the necessary and sufficient means for intelligent action" \cite{intelligence2003modern}. Their work culminated in the creation of many impressive AI systems including the creation of the Soar cognitive architecture \cite{laird1987soar}. 

\begin{figure}[h]
	\centering
	\includegraphics[width=0.85\linewidth]{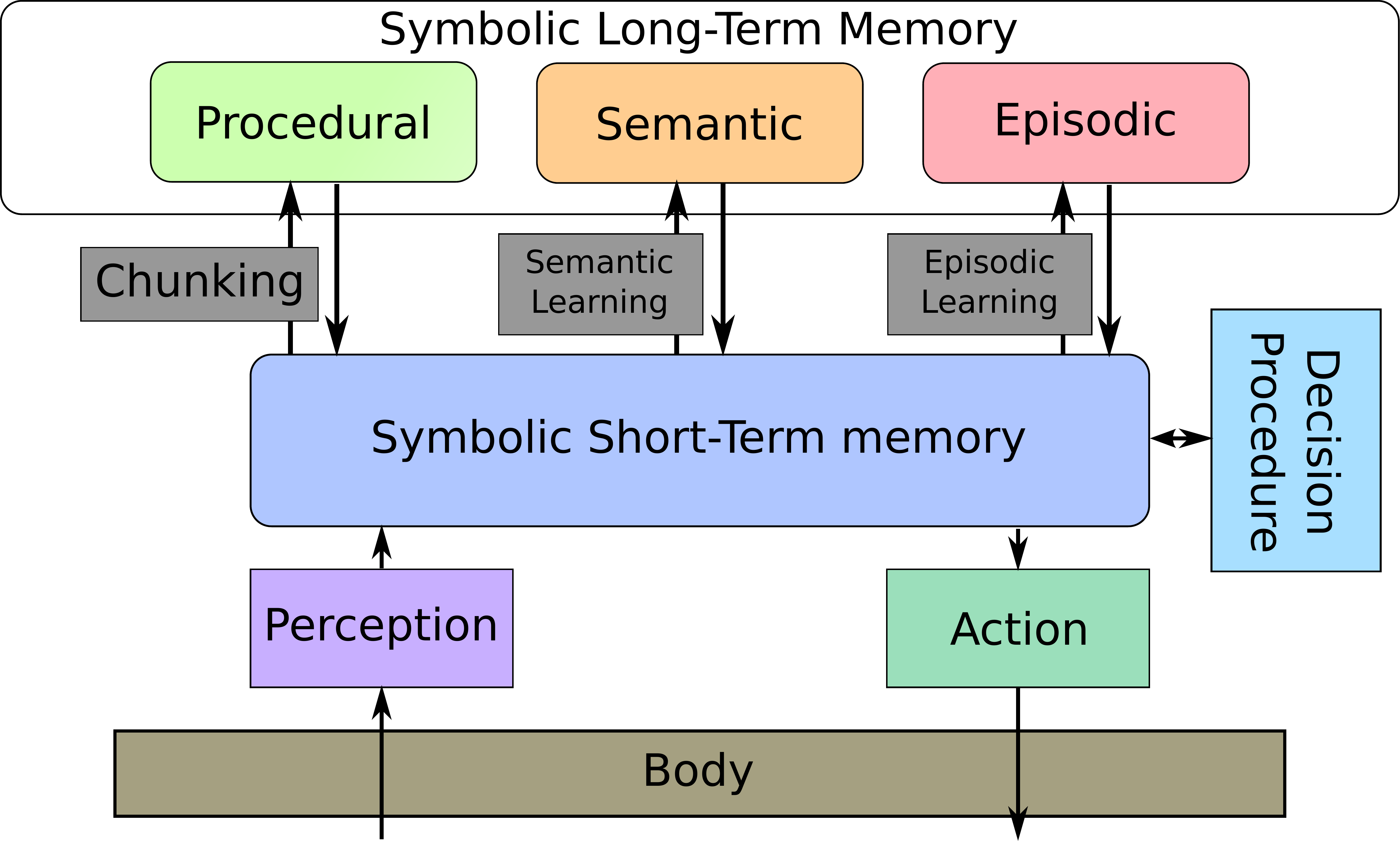}
	\caption{A simplified block diagram of the Soar cognitive architecture. \cite{laird2008extending}}
	\label{fig:Soar_block}
\end{figure}

Soar is one of many cognitive architectures that aims to create a formal, structured model of a cognitive system \cite{langley2009cognitive}. Figure \ref{fig:Soar_block} is an illustration of how Soar is composed and should clarify what a cognitive architecture entails. 

Formal structure plays an important role not only in Soar but all cognitive architectures as they must define features that remain constant in a cognitive agent \cite{langley2009cognitive}. The necessary components for cognition are defined by various cognitive models created by cognitive scientists. These components are designed as individual modules that can be interconnected to form a complete architecture. These models are typically based on the physical symbol system hypothesis, however there has been a shift towards hybrid-like approaches such as the CLARION \cite{langley2009cognitive} and Sigma \cite{rosenbloom2013sigma} architectures.

Cognitive architectures aim to utilise the knowledge contained in each of the different modules in a coherent and unified manner to produce cognitive behaviour \cite{langley2009cognitive}. The benefit of taking this approach is that agents can be designed with specific features that are well defined and can be understood by people. This is particularly important for high level cognitive capabilities that are easier to understand at a symbolic level than at a sub-symbolic level \cite{smolensky1990tensor}. From a scientific and engineering point of view this is an important attribute that is missing in connectionist models because they are so complex that they become incomprehensible for human interpretation (see Section \ref{sec:connectionist}). 

Autonomous vehicles are a popular application of AI in engineering and even though there has been great success in recent years the technology used still lack the capabilities most associated with intelligence such as problem solving and decision making \cite{hanford2011cognitive}. These capabilities are useful for machines operating in dynamic, unknown and uncertain environments and cognitive architectures bring these capabilities to the engineering community.  

A particularly interesting feature of humans is the ability to recall sequences of historical events that can be applied to the current situation rather than having to perform a new process to achieve the same outcome. This type of memory, known as episodic memory, has been introduced in Soar and it offers some impressive capabilities as shown by Nuxoll and Laird \cite{nuxoll2012enhancing}. This ability to retain sequences of actions and events could be useful in mobile robot navigation tasks where navigation becomes a recollection of movements and not an entirely new navigation process (which is often computationally expensive). 

Episodic memory is just one example of many high-level cognitive capabilities where the classical approach to AI has been at the forefront \cite{langley2009cognitive}. Cognitive architectures allow for easier implementation of the complex structures required for high-level cognitive capabilities compared to non-symbolic approaches that obscure these formal structures. While it is definitely possible for connectionist models to implement these exact structures (this must be true because the human brain is a biological connectionist system) the techniques and methods in achieving this seem out of reach for now \cite{haykin2009neural}.         
           
There are many cognitive architectures in development which makes it overwhelming to choose one to focus on. Additionally there is a steep learning curve required to use each of them proficiently, which is not helped by the lack of learning resources. Where resources are available they are often limited to ``toy" examples or are outdated. This presents a stumbling block for development especially when connectionist models have highly active communities and many resources with real-world examples that makes it easier to get involved. 

These issues may seem trivial in the broader scheme of things, however, they underline one of the major downfalls of using these approaches - rigidity.  The formal structure of cognitive architectures confines designers to specific tools and methods, whereas connectionist approaches follow the same guiding principles. Cognitive architectures are also reliant on humans to encode much of the necessary knowledge which creates many practical and theoretical problems \cite{HARNAD1990335} that will not be discussed in this paper.    

There is certainly a place for symbol system approaches in equipping machines with high-level cognitive capabilities. Symbolic AI systems also offer the advantage of providing insight and understanding that can guide cognitive machine design. Unfortunately this is provided at the expense of requiring greater human effort and more rigid structures. The connectionist views as explained in Section \ref{sec:connectionist} allow for greater autonomy which results in less human effort and a more flexible structure.       

\section{Sub-symbolic Approaches}\label{sec:connectionist}
\begin{figure}[h]
\centering
\includegraphics[width=0.8\linewidth]{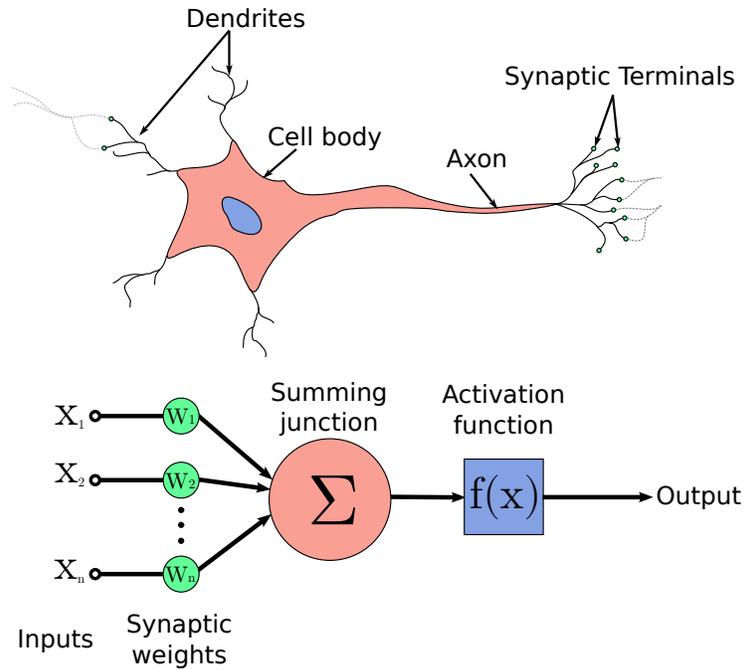}
\caption{An artificial neuron mimics the structure of a biological neuron. Neurons take in multiple weighted inputs, add them together and pass them through an activation function that determines the output.}
\label{fig:neurons}
\end{figure}

An alternative to symbolic AI is the connectionist approach that does away with formal processing blocks that model cognition in favour of an approach inspired by neurobiology \cite{smolensky1987}. Instead of relying on hand-engineered features and symbolic data structures connectionist models, such as artificial neural networks (ANN's), rely on the processing power of having many simple, interconnected processing units that allow for massively parallel computing \cite{haykin2009neural}. An illustration of the analogy between the artificial and biological neuron is provided in Figure \ref{fig:neurons}. 

These elementary building blocks allow ANN's to embed information in all of the weights across the network. An important feature is that the activation functions are mostly non-linear and are therefore useful for solving non-linear problems, whereas traditional approaches often involve the linearisation of problems so that linear techniques can be applied

The information contained in a neural network is a result of learning and not through direct encoding by a human \cite{smolensky1987}. Learning is often performed in a supervised manner (although unsupervised learning is improving) where the neural network is provided with an input and a desired (labelled) output. The input data is passed through the network to produce the actual output which can be compared to the desired output \cite{haykin2009neural}. Various algorithms exist to perform this comparison and use this to update the weights and biases in the network. Through appropriate training it is possible to create a network that can perform a desired task with little human involvement. Unlike the rigid structure of cognitive architectures, connectionist models are adaptive by nature \cite{haykin2009neural}.

The beauty of ANN's is that one can use them to accomplish many different tasks even though the principles and methods will remain the same \cite{haykin2009neural}. So whether one is interpreting radio data, extracting features from range finding sensors or even performing image compression, the fundamental principles remain the same. From a practical perspective this flexibility extends to how the same models can be built using completely different programming languages and hardware, whereas each cognitive architecture has their own rules for creating models and are mostly restricted to CPUs. The development of neuromorphic hardware as described in Section \ref{sec:neuromorphic} makes these models even more attractive. 

\begin{figure}
\centering
\includegraphics[width=0.75\linewidth]{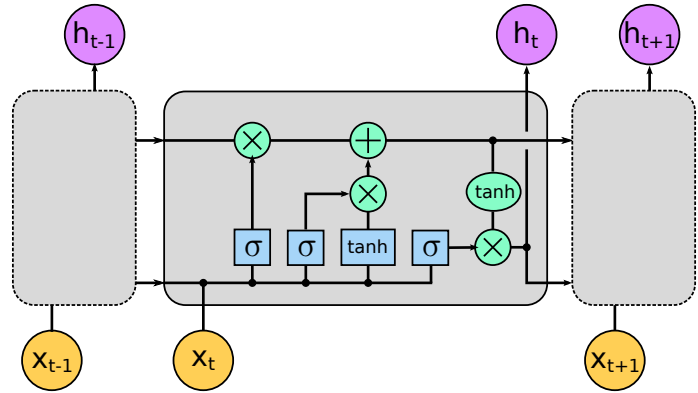}
\caption{This diagram illustrates how an LSTM memory cell can be``unfolded" over time. Where $x_t$ indicates the current input, and $h_t$ indicates the cell's current state. There are also various activation blocks that control what information gets stored in the cell (input gate) and when information is accessed from memory (output gate). \cite{Olah2015}}
\label{fig:LSTM_architecture}
\end{figure}

Recent developments have seen an increase in the number of neural networks with recurrent connections that essentially act as some form of memory \cite{lipton2015critical}. Instead of only seeing a snapshot of data at each time step the network is capable of using the data from previous time steps to assist in processing the current data \cite{lipton2015critical}. Training of these recurrent neural networks (RNN's) posed issues for many years, however new techniques have helped solve this \cite{hochreiter1997long}. For example a popular RNN model and learning algorithm known as long short-term memory (LSTM) introduces specialised gates that control the flow of information to allow for the learning of long-term dependencies \cite{hochreiter1997long}. In this model the basic building block is no longer just a neuron, but rather what is known as a ``memory cell" \cite{hochreiter1997long}. Figure \ref{fig:LSTM_architecture} illustrates the complexity of these models and hopefully reveals the difficulties associated with using them.

This increased complexity in RNN's makes them capable of learning more advanced features at the expense of becoming more difficult to train and compute \cite{Schmidhuber2015DLReview}. These extended capabilities are particularly attractive for engineering applications that rely on sequences of data such as interpreting motions recorded by an inertial measurement unit or range sensor data from a mobile robot.   

A major stumbling block in creating large neural networks, however, is the sheer quantity of model parameters that make it impossible for a human to comprehend \cite{smolensky1987}. This leads to difficulty in predicting the outcome of a given network. Other popular black-box models, such as transfer functions, differ from connectionist models in that there are well defined methods for analysing them. It is easier to predict how the model will behave and what changes need to be made to obtain the desired performance. Unfortunately connectionist models do not have such methods yet. This is problematic because engineering design concerns itself with understanding how design choices affect the performance of the system. 

This approach has worked well for perceptual tasks such as classification and recognition of patterns, however, this is a small piece of the cognition and there is still difficulty in learning complex representations necessary for high-level cognitive functions \cite{sun1999artificial}. Connectionist models have the potential to match and even enhance the capabilities of pure symbol systems, but these developments are likely still far away. 

A problem with having to compute large connectionist networks is that they require specialised hardware to compute efficiently. GPUs are better suited to the massively parallel computation required in neural networks compared to CPUs, however their power usage is still far beyond that of the human brain \cite{haykin2009neural}. This is not a problem for applications that can run on desktop computers or servers where there is sufficient power. It becomes an issue when creating mobile devices where power is limited. Alternative computing hardware could solve this, as discussed in Section \ref{sec:neuromorphic}.    

Having explored the two different approaches to AI it is suggested that a third option be looked at - a hybrid approach. \cite{HARNAD1990335} suggests that this is a viable and attractive solution to many problems that each individual approach is faced with and that it may be necessary to accomplish cognitive machines.       

\section{Hybrid Approaches}\label{sec:hybrid}

As mentioned before, symbolic and connectionist models are each suited to specific levels of cognitive capabilities. Symbolic models are able to perform high-level cognitive tasks such as reasoning and planning while lacking sufficient capabilities of handling low-level tasks such as perception and action. Connectionist approaches have been extremely successful in perceptual tasks and are useful in adaptive control but so far lack the high-level capabilities that are necessary for complex tasks \cite{HARNAD1990335, smolensky1990tensor}. 

In the current applications that have made neural networks popular, such as image recognition, there has been no real need for high-level capabilities. In the case of machines such robots, there is a need for strong perception and action capabilities because the agents must interact with the physical world \cite{haykin2012cognitive} and there is also the need to include problem solving and decision making capabilities for the robotic agents to operate without human intervention \cite{haykin2012cognitive}. Various industries already employ robots to minimize the need for humans to perform dangerous tasks or tasks that require extremely high precision and accuracy. Some applications have been out of reach due to the lack of high-level cognitive capabilities.  

It is possible to bring cognitive machines closer to realisation by combining the strengths of symbol systems with connectionist models, as well as other non-symbolic approaches. Hybrid cognitive robotic architectures have been explored before in \cite{kelley2009hybrid} and \cite{hanford2011cognitive} but there remains a wealth of untapped capabilities such as the use of episodic memory. 

The SS-RICS architecture in \cite{kelley2009hybrid} used a common robotics approach for generating a map for navigation that utilizes metric information from sensor data. They encountered various issues with this in that the classification of intersections based on sensor data was often incorrect and compounded as the robot continued its task. They argue that without a useful perceptual system the higher-level capabilities can never be realised because there would be difficulty in creating meaningful symbolic relationships \cite{kelley2009hybrid}. The CRS architecture used in \cite{hanford2011cognitive} used fuzzy logic to improve the classification of intersections but odometery errors meant that the robot mistakenly identified the same intersections as different ones. 

In both SS-RICS and the CRS architectures the majority of the faults were with the perceptual systems used. Despite this, both attempts showed some useful results from their experiments that showed a glimpse of what could be possible with a hybrid system should the perceptual systems have been up to the task.

Thankfully perceptual systems have improved substantially as mentioned in Section \ref{sec:connectionist}. An example of where a cognitive machine could leverage a hybrid cognitive architecture is in robotic mapping. Traditional mapping techniques such as building occupancy grids caused issues in SS-RICS and the CRS as mentioned above. Even though they both mention that such techniques are not cognitive processes they continue to use them as a step towards providing semantic labels for a high-level symbol system \cite{kelley2009hybrid, hanford2011cognitive}. 

\begin{figure}[h]
\centering
\includegraphics[width=0.55\linewidth]{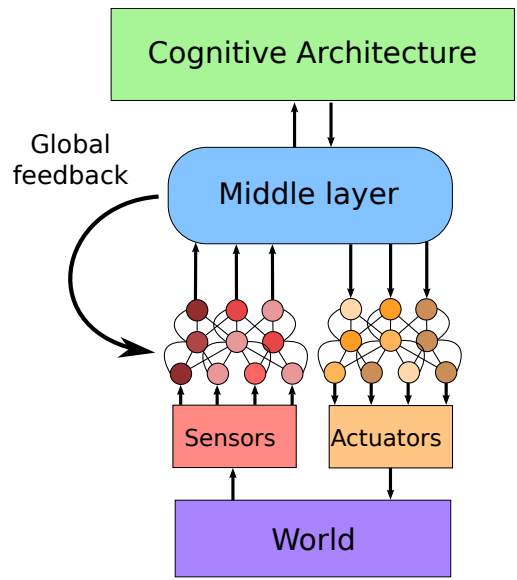}
\caption{A cognitive architecture can be used as a deliberative layer because of its high-level cognitive capabilities. The middle executive layer would control the flow of information between the deliberative and reactive layer. An ANN would make an appropriate reactive layer for perception and action.}
\label{fig:hybrid_hierarchy}
\end{figure}

This paper proposes a hybrid architecture intended for use in a mobile robot that can be realised by combining the different approaches in a hierarchical fashion as shown in Figure \ref{fig:hybrid_hierarchy}.

The proposed architecture could be applied to robot navigation where it could use RNNs to utilize sequential sensor data to construct useful local representations of the local environment. The advantage of using ANNs is that they are capable of extracting better features than those that are hand-engineered. The cognitive architecture could use that local information to construct topological maps which allow for easier path planning, decision making and problem solving compared to metric maps \cite{hanford2011cognitive}. Topological maps are also far more compact and will have less memory requirements. There could also be a global feedback loop that can augment the perceptual system for enhanced capabilities i.e. it may be possible to use the cognitive architecture to control what features the perceptual system should focus on. 

The tight integration between the different layers is something that will need to be looked at carefully. The hybrid model needs to be designed in such a way that the addition of a structured symbol system does not inhibit the flexible nature of the connectionist system and that the connectionist system is capable of forming meaningful symbolic relationships.     

\section{Neuromorphic Emulation}\label{sec:neuromorphic}

\begin{figure}[h]
\centering
\includegraphics[width=0.75\linewidth]{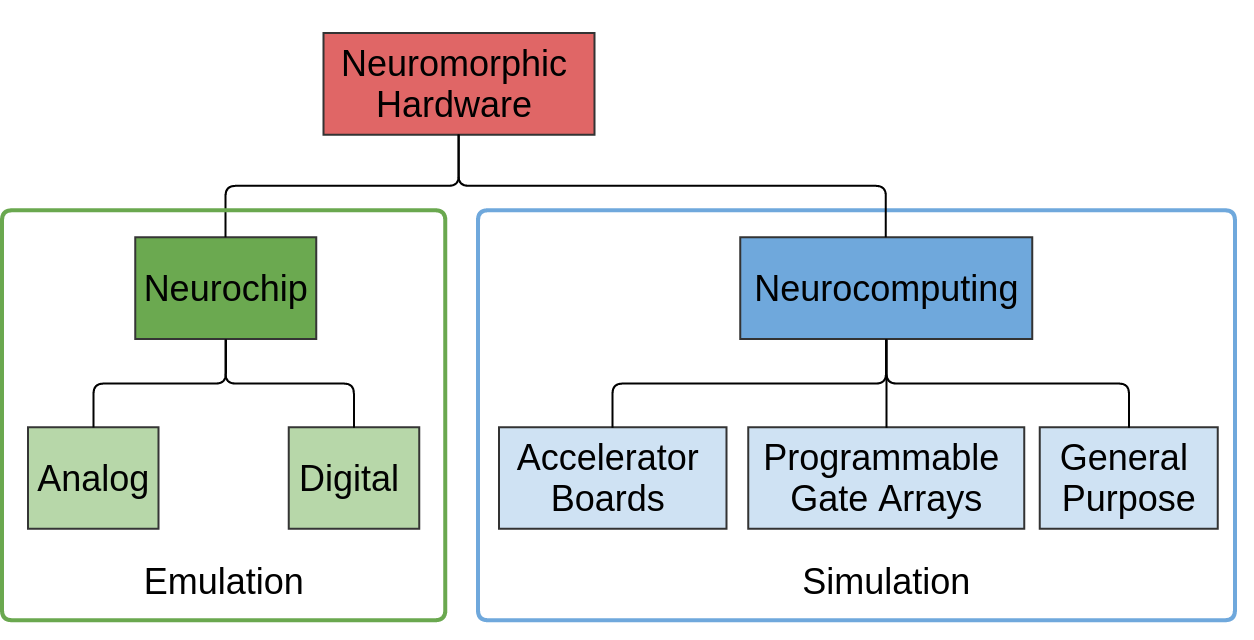}
\caption{Neuromorphic hardware can be divided into emulations and simulations that describe in what manner the neural networks are implemented.}
\label{fig:Neuromophic_hardware_tree}
\end{figure}

As mentioned before in Section \ref{sec:connectionist} there is a need for specialised hardware to implement connectionist models. The Human Brain Project has a platform that aims to emulate the functioning  inside the human brain. They provide a review of neuromorphic technology in \cite{calimera2013human} where they provide a breakdown of neuromorphic hardware as shown in Figure  \ref{fig:Neuromophic_hardware_tree}.

Simulating large scale neural networks using von Neumann architectures is inefficient and would require incredibly large amounts of power \cite{benjamin2014neurogrid, merolla2014million, furber2014spinnaker}. One of the primary bottlenecks to overcome is the inefficient movement of data that occurs between processors and memory in traditional von Neumann architectures \cite{merolla2014million}. In order to emulate neural networks there needs to be a tighter integration between processors and memory to form the individual neurons. Various approaches to this idea have been undertaken by research teams from around the world. Most notable are the works done by the University of Manchester with their SpiNNaker project \cite{furber2014spinnaker}, IBM with their TrueNorth architecture\cite{merolla2011digital}, Stanford University and their Neurogrid architecture \cite{benjamin2014neurogrid}, and a team from  the University of California at Santa Barbara \cite{prezioso2015training}.

\begin{figure}[h]
\centering
\includegraphics[width=0.4\linewidth]{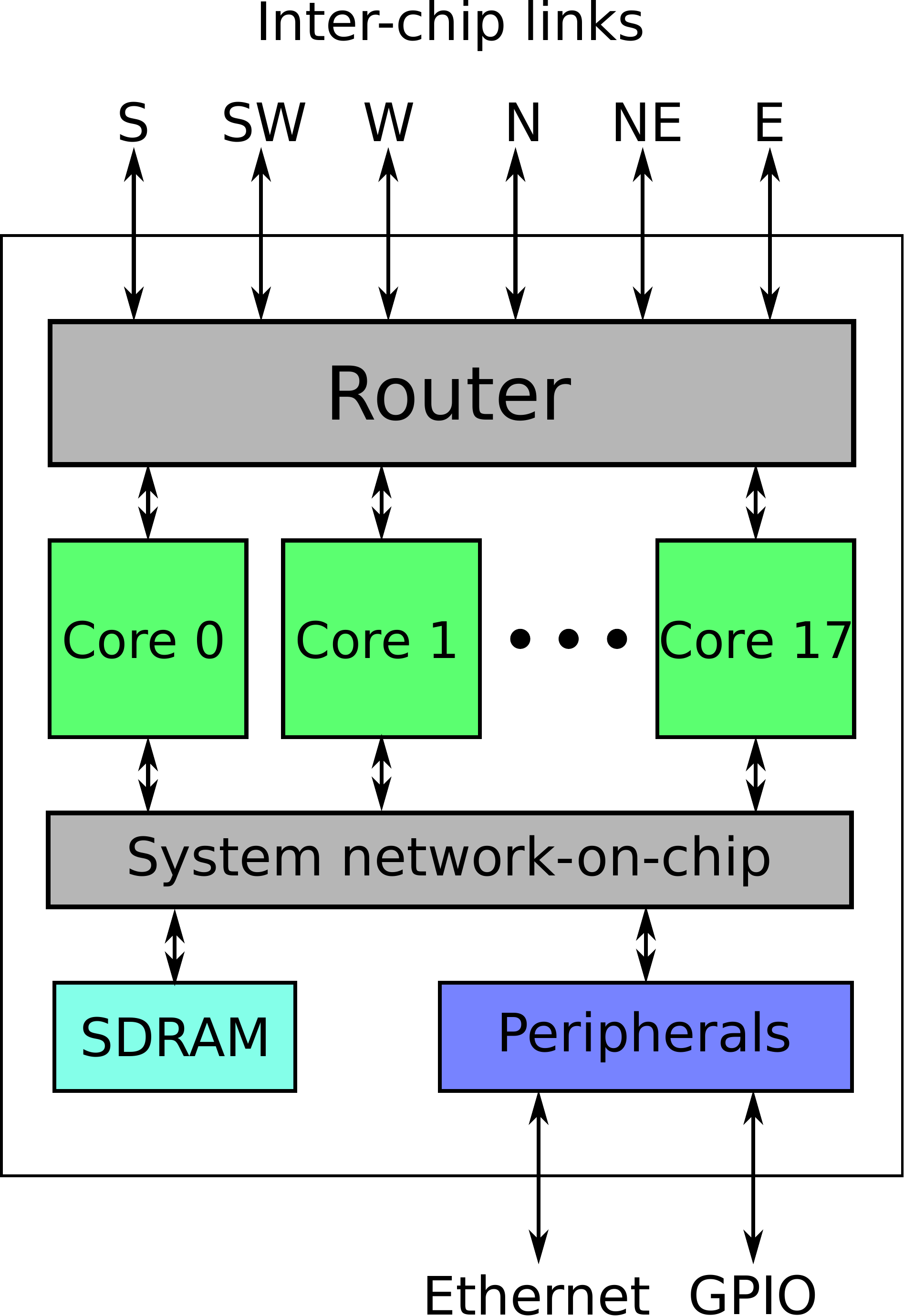}
\caption{A SpiNNaker node consists of interconnected blocks of existing digital components. \cite{furber2014spinnaker}}
\label{fig:SpiNNaker_Node}
\end{figure}

The University of Manchester have done work on designing and implementing a neurally inspired computational hardware as part of the Neural Computing Platform for the Human Brain Project. The architecture employed in their SpiNNaker project utilizes processing nodes consisting of 18 general purpose ARM968 cores and extra memory for each node \cite{furber2014spinnaker}. The block diagram for one node is provided in Figure \ref{fig:SpiNNaker_Node}.

They then place multiple processing nodes on a single PCB with FPGA's for high-speed interconnectivity between the nodes. They have adopted a simplistic approach to achieving massively parallel computing by opting to use large quantities of existing processors rather than designing custom circuitry to emulate individual neurons \cite{furber2013overview}. The use of existing technology allows for quicker prototyping and construction (especially when considering the scale to which they are aiming to achieve). Much of the alternative research involves the design and implementation of custom circuitry to emulate neurons more closely to achieve better efficiency.  

\begin{figure}[h]
\centering
\includegraphics[width=0.7\linewidth]{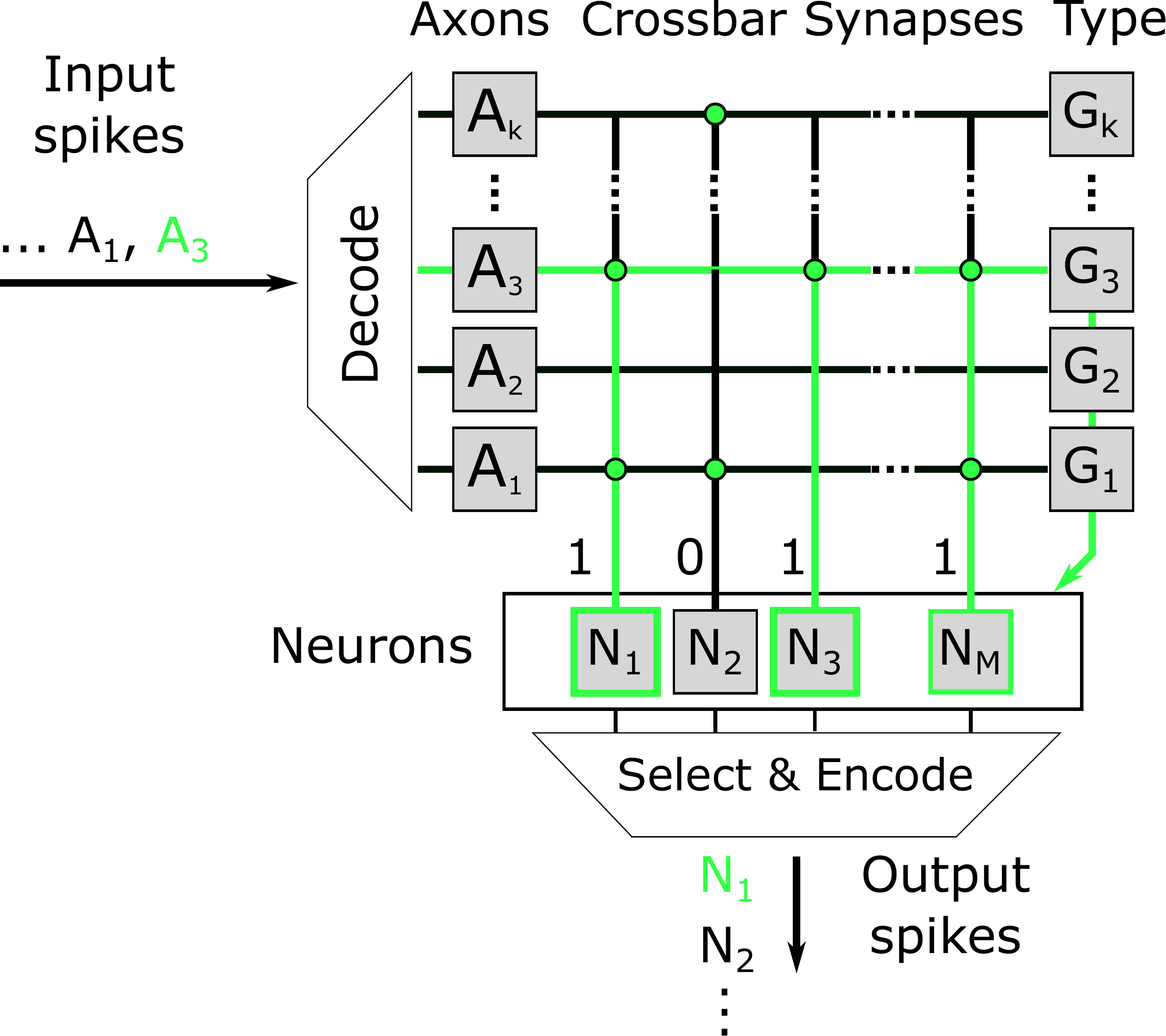}
\caption{The TrueNorth architecture utilizes a custom arrangement of digital circuits to emulate a mesh of neurons. \cite{merolla2011digital}}
\label{fig:Neurosynaptic_core}
\end{figure}

At IBM they have developed what they call digital neurosynaptic cores as the fundamental building blocks of their TrueNorth architecture \cite{merolla2011digital, merolla2014million}. They are able to implement spiking neural networks by using existing digital electronics ``blocks" such as decoders, encoders and SRAM in a mesh structure to emulate the axons, neurons and synapses respectively. The structure of their implementation is shown in Figure \ref{fig:Neurosynaptic_core}.
This approach does not use existing computer architectures such as the ARM cores used in the SpiNNaker architecture. The benefit of this is an increased number of neurons and synapses per chip as well as improved efficiency. TrueNorth has 1 million neurons and 256 million synapses \cite{merolla2014million} compared to SpiNNaker's approximate number of 18 thousand neurons and 18 million synapses per node \cite{furber2013overview}. Despite this achievement there are still further developments in replicating the efficiency of neurons by delving into analog electronics.

\begin{figure}[h]
\centering
\includegraphics[width=0.6\linewidth]{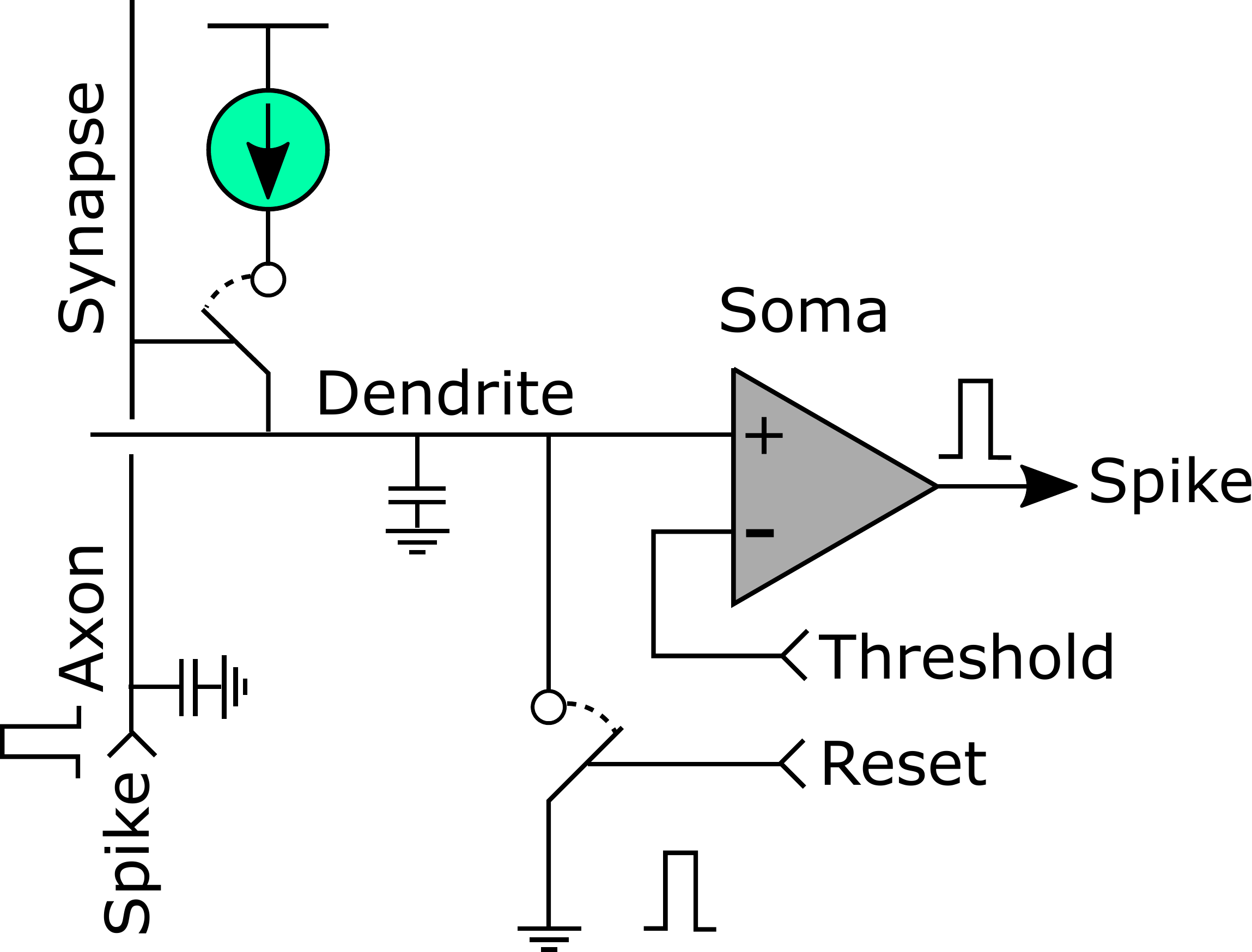}
\caption{Analog elements are able to directly emulate the function of a biological neuron from physical principles. \cite{benjamin2014neurogrid}}
\label{fig:analogneuron}
\end{figure}

The team working at Stanford are working on a mixed analog-digital hardware platform for neural computing called Neurogrid \cite{benjamin2014neurogrid}. Their aim is to reduce the power requirements for neural computing as much as possible by using sub-threshold analog electronics. Rather than relying on digital memory to store synaptic weights the Neurogrid allows for these values to be stored directly in the electronic make-up of the neurons. An example of an analog silicon neuron is provided in Figure \ref{fig:analogneuron}.
Their technology has proved to be very efficient \cite{benjamin2014neurogrid} and is a highly promising project. 

At the University of California at Santa Barbara they have been working on creating neural networks that use the well suited properties of memristors \cite{prezioso2015training}. Memristors have a fundamental property that is very much like the synaptic connection between neurons \cite{thomas2013memristor}. Where an increase in flux in one direction causes the resistance to increase and flux in the opposite direction causes resistance to decrease \cite{thomas2013memristor}. In their paper \cite{prezioso2015training} they were able to train a single-layer perceptron network to classify a 3x3-pixel image without the use of any CMOS components - only memristors. The simplicity of the circuit makes it a very intriguing prospect and certainly a more accurate model of a neuron. 

Each of the mentioned projects were chosen to provide an overview of the range of approaches one could use to design neurmorphic chips. The technology is no doubt still in the early stages of development but it is clear that cognitive machines will rely on such technology in the future.

\section{Conclusion}

This paper was intended to provide a brief review of technologies that can assist in enabling the creation of cognitive machines. In the relatively brief history of AI much has changed over the years due to new scientific insights and rapid technological growth. The symbolic AI systems that excelled early on were stunted by the oversimplification of cognitive mechanisms \cite{intelligence2003modern}. In an opposite trend the early abandonment of connectionist models has been reversed in astounding fashion due to technological advancements that have made massively parallel computation more feasible \cite{intelligence2003modern}. 

A hybrid cognitive architecture utilizing state-of-the-art techniques from both approaches is a viable option for creating machines that are enhanced with cognitive capabilities. Prior attempts at creating hybrid approaches have not integrated the absolute best of both worlds. As mentioned by Smolensky \cite{smolensky1990tensor} it is ill-advised for the two camps (symbolic and connectionist) to ignore each other, however major developments in hybrid models have fallen behind in comparison to developments in the individual fields. While the ultimate goal may be to have a full hardware realization of a neural network, a hybrid cognitive model may allow for sufficient capabilities to outperform existing machines in the mean time. The technological landscape is changing and cognitive engineering is very much at the forefront.

\bibliographystyle{IEEEtran}
\bibliography{ref}

\end{document}